\documentclass[letterpaper, 10 pt, conference]{ieeeconf}  % Comment this line out if you need a4paper

\IEEEoverridecommandlockouts% This command is only needed if 
\usepackage{subfig}
\usepackage{graphicx}
\usepackage{comment}
\usepackage{booktabs}
\usepackage{diagbox}
\usepackage{multirow}
\usepackage{colortbl}
\usepackage{makecell}
\usepackage{xcolor}
\usepackage{cite}
\usepackage{url}

\usepackage{amsmath} % assumes amsmath package installed
\usepackage[colorinlistoftodos]{todonotes}

\title{\LARGE \bf
A Perception-Manipulation Robotics System for Food Cutting
}

\author{Xinyuan Luo$^{1}$, Wenzhen Yuan$^{1}$% <-this % stops a space
\thanks{$^{1}$Xinyuan Luo and Wenzhen Yuan are with the University of Illinois at Urbana-Champaign \{\tt\small xl153, yuanwz\}@illinois.edu}%
}

\begin{document}
\maketitle
\thispagestyle{empty}
\pagestyle{empty}

%%%%%%%%%%%%%%%%%%%%%%%%%%%%%%%%%%%%%%%%%%%%%%%%%%%%%%%%%%%%%%%%%%%%%%%%%%%%%%%%
\begin{abstract}
% \wenzhen{Now the paper exceeds the page limit}
In the development of cooking robots, mastering the task of cutting is crucial. A significant challenge lies in the diverse properties of food, which necessitate distinct cutting policies and even different knives for optimal processing. This paper presents a perception-manipulation framework for food-cutting tasks. Our system features a knife selection module that utilizes force data from a preliminary fixed trial cut to select the appropriate knife for the given food. This is followed by an adaptive cutting phase using reinforcement learning (RL) to balance cutting speed and energy efficiency. In our experiments, the knife selection module achieved 100\% successful rate on unseen food, and we compared the performances of fixed policy, RL policy, with human operators. Our method not only achieves high performance but also demonstrates comparable results to those of human participants. Project page: \url{https://robotproject8.github.io/robocutting/}
% \amin{"which optimizes the cutting process for energy efficiency" I guess you can rewrite this sentence. optimization and efficiency are similar I feel, and is repeating}
\end{abstract}

%%%%%%%%%%%%%%%%%%%%%%%%%%%%%%%%%%%%%%%%%%%%%%%%%%%%%%%%%%%%%%%%%%%%%%%%%%%%%%%%
\section{Introduction}

% cooking robot intro
The development of cooking robots has been a key interest among researchers and industry practitioners for many years. The comprehensive cooking process includes food preparation tasks such as cutting \cite{deepMPC}, peeling \cite{ye2024morpheusmultimodalonearmedrobotassisted}, and stirring \cite{luo2024intelligentroboticperceptivepancake}, along with various cooking techniques like flipping \cite{roboFlip} and stir-frying \cite{stir-fry}. Most of these sub-tasks present significant challenges due to the diverse properties and irregular shapes of food, as well as the delicate manipulation skills required.

% mini intro of related works
Cutting, a critical process in food preparation, necessitates a thorough understanding of food properties such as hardness and friction coefficient. While humans have an innate sense of these properties before cutting, they also adapt their cutting techniques in real-time based on the force-torque feedback received from knife contact. This adaptation control can be achieved by learning a dynamic model for model predictive control (MPC) \cite{deepMPC,data-drivenMPC,lstmMPC}, employing preset control algorithms \cite{control1,control2,control3}, and utilizing reinforcement learning (RL) \cite{RL1, rl2}.
% \amin{the last sentence was not clear to me }

\begin{figure}[htbp]
\begin{center}
\includegraphics[scale=0.65]{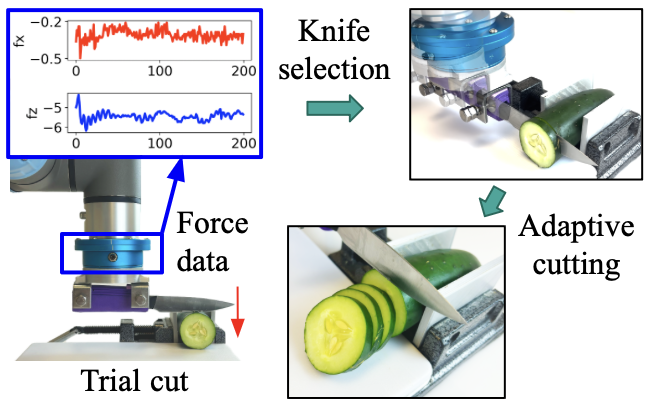}
\end{center}
\caption{\small We developed a food cutting system that first performs a fixed trial cut motion to the food to understand the food properties and select a proper knife. Following this, the food is cut using an RL adaptive controller. }
% \wenzhen{The F/T should labeled at the wrist. The caption of this figure doesn't make sense. also the "trial cut" seems weird -- it's not really touching the food}
\label{fig:teaser}
\end{figure}

\begin{figure*}[htbp]
\begin{center}
\includegraphics[scale=0.55]{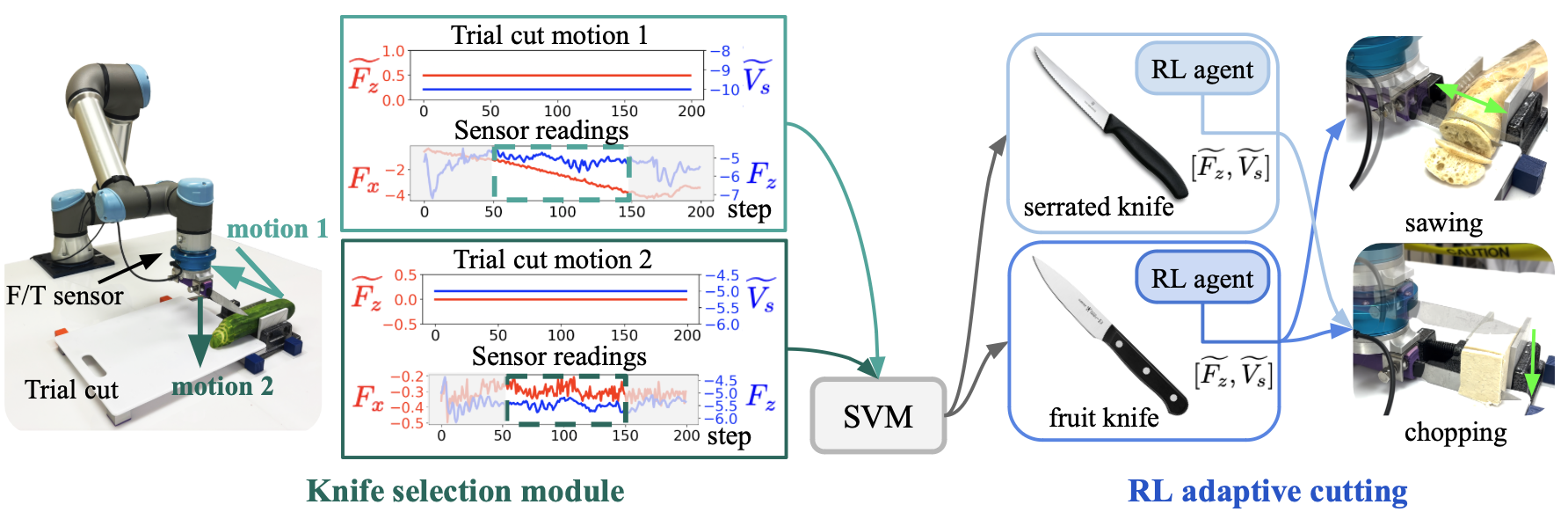}
\end{center}
\caption{\small Pipeline of our Perception-Manipulation Robot Cutting System for an arbitrary food item: the robot first conducts two trial cut motions (defined as fixed target vertical force $\widetilde{F_z}$ and a fixed sawing speed $\widetilde{V_s}$) and uses the measured force value $F_x$ and $F_z$ to decide the type of knife to use. Subsequently, the robot with the selected knife, will execute cutting tasks using the policy $[\widetilde{F_z}, \widetilde{V_s}]$ output by the pre-trained reinforcement learning (RL) agent.
% : When the robot encounters a new food item, it first conducts trial cuts of sawing (motion 1) and chopping (motion 2). Utilizing the F/T sensor, the $f_x$ and $f_z$ series during the trial cuts are sampled and put into an SVM to classify which knife to use. Subsequently, a corresponding pre-trained model is loaded to execute RL-based adaptive cutting.
% \wenzhen{In the leftmost figure, the labelings of two motions are not recognizable. you can probably just remove them; you need a small title for the Fz Fx plot saying that the plots are "sensor reading"; it's best to change the label to "target fz"; no x-axis label in the figures; might be nice to put two set of plots in two boxes; there is nowhere you showed the RL policy, it's good to add a box between the knife and the rightmost figure indicating that you are running RL}
}
\label{fig: system pipeline}
\vspace{-5mm}
\end{figure*}

% in this work general
In this work, we aim to develop a food-cutting system that generalizes to various types of unseen foods. 
% \wenzhen{This is not a good enough summarization. You should write in high level what the method is about -- don'e make any assumption that the readers are familiar with your problem setting and method. Also here you need to provide some insight about the problem itself, and why you choose this method}
Similar to humans, our robot initially interacts with the food to assess its properties. During the cutting process, the robot can dynamically adjust its cutting strategy based on force feedback. Fig.~\ref{fig:teaser} provides an overview of the proposed system. To expand the variety of foods it can handle, we have equipped the robot with an easy-to-switch knife mount that allows seamless transitions between a fruit knife and a serrated knife. As shown in Fig \ref{fig: system pipeline}, the whole process begins with a fixed trial cut to collect force data, which is then used to determine the appropriate knife. Then, we have designed a reward system to measure cutting efficiency and employed an on-policy RL agent to online refine the cutting motions.

% Once the cutting process is complete, the system can retain the cutting model, enabling it to recognize and handle similar types of food in the future effectively.

% \amin{you have mentioned the knife switch couple of times}
% challenge 1, our solution
A core challenge in food cutting stems from the diverse properties of food, such as hardness, surface friction coefficient, and juiciness. For instance, 
% while both chopping and sawing can effectively cut through a cucumber, 
chopping is unsuitable for baguette due to its deformable nature; instead, a serrated knife and a sawing motion are required. Additionally, cutting juicy foods like tomatoes too hard may lead to damage due to their delicate structure \cite{LeTac-MPC, learnByBreak}. To tackle this challenge, we designed an easy-to-switch knife mount shown in Fig \ref{fig: setup}(c) that could effectively switch between a fruit knife and a serrated knife. Utilizing the information obtained from the trial cut, the system can choose a proper knife and perform sawing or chopping. % challenge 2, our solution

% \amin{I feel the next paragraph is too detailed for intro}
Another significant challenge is the convergence of the adaptive reinforcement learning controller. 
% Since our system does not utilize simulations, conducting extensive real-world training could result in considerable food waste. 
To expedite the convergence of the RL algorithm, we carefully design the state space, action space, and reward function. We then develop an initial value selector to accelerate the convergence.
% utilize the critic network of this model to choose the best initial parameters for the adaptive controller. 
These approaches enable our system to achieve convergence within the cutting of just $2$ to $3$ slices.

% The process is that we first undergoes a perception phase that utilizing multi-model perception includes tactile, force-torque from a trial cut and vision to choose proper knife and model to cut the food. Then a corresponding pre-trained model is loaded to cut the food into slices. This model is online adaptive according to the reward during the cutting process and optimized by an RL algorithm. After the food cutting is done, the newest model will be saved along with the representation of the food properties obtained from the perception phase. When the next time encountered a new food, the system will compared the similarity of the representation and choose a most similar model to it. Thus is able to improve it's cutting skills with cutting more and more food.

% contribution summary
% In summary, this work's main contributions include 1) A trial cut module to select the proper knife and cutting policy and achieve $100\%$ accuracy for $4$ kinds of unseen food. 2) An adaptive controller optimized through real-world reinforcement learning, capable of quickly generalizing to new food types and achieves better performance than fixed policy and comparable performance to human. 3) A comprehensive robotic cutting pipeline that incorporates knife/policy selection and adaptive cutting.
In summary, the key contributions of this work are 1) A trial cut motion that selects the appropriate knife and cutting policy, achieving $100\%$ accuracy with $4$ types of unseen food. 2) An adaptive controller that is optimized via real-world reinforcement learning, capable of rapidly generalizing to new types of food and outperforming fixed policies while delivering results comparable to human performance. 3) A comprehensive robotic cutting pipeline that integrates knife and policy selection with adaptive cutting.

% paper organization
% \wenzhen{This paragraph is not important. However, in the introduction, you should talk about what result you got and how do you understand the results}
% This paper is structured as follows: Section \ref{sec: related works} reviews related work on cutting robots and multi-modal perception of food properties. Section \ref{sec: method} details the multi-modal perception phase and the adaptive RL controller. Section \ref{sec: experiments} describes the experimental setup and neural network parameters. In Section \ref{sec: experiment result}, we compare our results with baseline performances and human cutting. The paper concludes by summarizing our contributions and discussing future directions for this research.

\section{Related Works}
\label{sec: related works}

\subsection{Robotics cutting}
% Other interesting cutting problems

Researchers tackle the cutting task using either control-based methods \cite{control1, control2, control3} or learning-based methods.

% $$
% \textcolor{red}{\widetilde{F_z}}
% $$
% $$\textcolor{blue}{\widetilde{V_s}}$$
% $$\textcolor{red}{F_x}$$
% $$\textcolor{blue}{F_z}$$
% $$\widetilde{F_z}$$
% $$\widetilde{V_s}$$

% MPC
Due to the significant effort required to tune control parameters for foods with varying properties, researchers have turned to model predictive control (MPC). Lenz et al. \cite{deepMPC} introduces DeepMPC, which employs a novel deep neural network structure to learn the dynamics of various foods. Similarly,  Mitsion et al. \cite{lstmMPC} utilizes an LSTM network to learn the intrinsic dynamics, and their another work \cite{data-drivenMPC} employs a comparable learning and control framework. However, these approaches generally require the collection of large datasets to accurately learn the dynamic models of various foods.

% DMP
Dynamic motion primitives (DMP) \cite{dmp} represent another approach to robotic cutting \cite{dmp_grapefruit}. Zhang et al. \cite{cutting-audio} employs DMP as a behavior cloning method to learn from human demonstrations of cutting various foods, and also to generalize to unseen food types. Yang et al. \cite{dmp_huanan} also incorporates cutting tasks into their experiments. 
% However, teaching robots to cut in an intuitive manner remains challenging.

% RL
Reinforcement learning offers promising directions for robotic cutting. Padalkar et al. \cite{RL1} employs Policy learning by Weighting Exploration with the Returns (PoWER), learning parameters for an impedance controller in downward motions combined with a fixed sawing motion to effectively cut cucumbers and bananas. Meanwhile, Beltran-Hernandez et al. \cite{rl2} leverages the DiSECt simulation engine \cite{rl_simulation}, specifically designed for cutting tasks. This approach models food as a spring, uses a trial cut to determine its parameters, and learns cutting skills through simulation before transferring them to a real robot. In contrast, our trial cut is not used to explicitly estimate physical parameters for simulation. Instead, we use force features from trial cuts to select the knife and initialize a real-robot cutting policy, and then use on-policy reinforcement learning to fine-tune the target vertical force and sawing speed during execution. This positions our method as a real-world perception-manipulation pipeline rather than a parameter-identification step for simulation-based training.

% \cite{cutting-audio} introduce a method that leverages vibrations and force-torque feedback to adapt slicing motions and monitor contact events, allowing a robot to cut through a diverse array of food items by learning neural networks that generalize across different material properties. \cite{cutting-MPC} explore the modeling and learning of dynamic interactions for robotic food-cutting, employing a data-driven approach with a Long Short-Term Memory (LSTM) model and MPC to approximate the intricate dynamics of food-cutting, leading to an enhanced understanding of cutting dynamics and generalization to unseen object classes. Further expanding the scope, another study introduces RoboNinja \cite{cutting-ninja}, which learns an adaptive cutting policy for multi-material objects, illustrating the potential of deep learning to navigate the complexities of cutting tasks involving materials with varying properties. However, the majority of cutting research focuses on cutting strategies and typically employs a fixed strategy to stabilize the object.

\subsection{Perception of food properties}

Conducting food manipulation tasks necessitates an understanding of food properties. Numerous researchers have explored this area. 
For example, Mu et al. \cite{property1}  propose a recursive least-squares method to estimate relevant physical parameters, such as Poisson’s ratio, fracture toughness, and the coefficient of friction, using only force sensor readings. Sawhney et al. \cite{playwithfood} introduce a multimodal sensory approach that combines vision, finger vision, and audio to learn embeddings that represent food type, juiciness, hardness, slice type, and width effectively. They also use multimodel perception to classify the current state of a cutting system, for example, the knife contacting the food or scraping the cutting board \cite{cutting-audio}. Similarly, Ye et al. \cite{ye2024morpheusmultimodalonearmedrobotassisted} utilizes vision, vibration, and force to determine whether food has been peeled. As a preliminary study to DeepMPC, Gemici et al. \cite{deepMPCproperty} investigates various actions using different tools to learn about food properties such as hardness, plasticity, elasticity, tensile strength, brittleness, and adhesiveness. Our method is related to these probing-based approaches, but it uses the trial interaction as a compact decision signal for knife selection and policy initialization, without requiring an explicit estimate of material parameters.

\section{Method}
\label{sec: method}
% \amin{explain the reason to choose speed and force?}
To address the challenge of adaptively cutting any unseen food, we constructed a low-level controller with two parameters: the target force in the vertical direction ($\widetilde{F_z}$)
% \wenzhen{I think it's better to use "target force". You can check with others. Also somewhere else you used Fz. It's a little confusing. You can use something like $\widetilde{F_z}$} 
and the sawing speed ($\widetilde{V_s}$). By tuning these parameters, the robot can perform chopping or sawing at various speeds, adapting to foods with different physical properties. To expand the variety of foods the robot can successfully cut, we developed a knife selection module that enables the robot to switch between a fruit knife and a serrated knife. To decide which knife to use, the robot performs a set of trial cuts with predefined motions to collect force data. This data is then used to solve a binary classification problem to select the appropriate knife. 
% Subsequently, we implemented an RL adaptive controller designed to balance cutting speed with energy efficiency. We designed a reward function that encourages the robot to cut down while being more energy efficient.
We then implemented an RL adaptive controller to balance cutting speed and energy efficiency, with a reward function that encourages efficient cutting.

\subsection{Low-level cutting controller}
\label{subsec: Low-level cutting controller}

% \begin{figure}[htbp]
% \begin{center}
% \includegraphics[scale=0.4]{figure/cut_config.png}
% \end{center}
% \caption{Cutting robot settings}
% \label{fig: cutting configuration}
% \end{figure}

% Fig. \ref{fig: cutting configuration} illustrates the configuration of our cutting robot settings. 
Both the trial and main cutting motions are controlled by the force ($\widetilde{F_z}$) in the z-direction (vertical) and the velocity ($\widetilde{V_s}$) in the x-direction (sawing). During the motion of cutting a single slice, the knife's movement is confined to the x-z plane. The knife is attached to a force/torque (F/T) sensor mounted at the robot's terminal. The food is fixed by a sliding vice. Fig \ref{fig: setup}(a) depicts the setup in a real-world scenario. 

On the z-axis, the knife is controlled by a PID force controller to apply down force. The control output, \( u(t) \), the output velocity of the robot terminal in vertical direction, is defined as: $u(t) = K_p \cdot e(t) + K_i \cdot \int e(t) \, dt + K_d \cdot \frac{d}{dt} e(t)$, where \( e(t) = F_z - \widetilde{F_z} \) defines the error between the measured force \( F_z \) and the target vertical force \( \widetilde{F_z} \). The parameters \( K_p \), \( K_i \), and \( K_d \) represent the proportional, integral, and derivative gains, respectively. 

% \amin{it would be better to provide more info on the position controller  }
On the x-axis, the sawing motion of each back-and-forth movement is governed by a position controller that tracks the commanded sawing speed, with the displacement constrained within a range of \(\pm 2\,\text{cm}\).

% \wenzhen{This paragraph is irrelevant here. You can talk about the set up of baseline method in the experiment session}
% The desired force $F_d$ and the sawing speed $V_s$ are two important parameters of this cutting policy. In baseline we choose two parameters manually and in RL adaptive controller, the RL agent will choose the parameters based on states in a frequency of $10$ Hz.

\subsection{Knife selection module}

% Why we need knife selection module
% \amin{what do you mean a regular blade?}
Intuitively, a fruit knife, with its flatter blade, typically produces flatter slices indicative of superior cutting quality; 
% hence, it is the default choice for achieving better results. 
However, we observed challenges or failure when using a fruit knife to cut hard foods like potatoes and sweet potatoes, as well as deformable and tough foods such as baguettes and other breads. 
% Attempts to train the RL controller to cut sweet potatoes and baguettes with a fruit knife frequently resulted in timeout failures. 
In contrast, a serrated knife more easily cuts such foods. Therefore, 
% to expand the range of foods our system can effectively cut, 
a knife selection module is essential.

% \xinyuan{What are those features? should be clarify after adding new exp result}
To classify food with those features and select a proper knife, we designed a trial cut to perform a pre-defined motion on food using the fruit knife. Utilizing force data obtained from the trial cut, we can decide whether to continuously use the fruit knife or switch to a serrated knife.
% using our specially designed easy-to-switch mount in Fig \ref{fig: setup}(c).
% \amin{I didnt understand what specific data you collect for classification and what was the reason for it?}
% parameter selection and evaluation
To find the best trial cut policy, we collected force, velocity, and displacement data across $14$ different foods using $6$ sets of trial cut parameters. Our analysis revealed that the force series data in sawing and vertical directions are the most informative. Experiments detailed in Section \ref{subsec: knife selection module} demonstrated that a combination of fixed sawing and chopping motions most effectively captures the properties of the food, resulting in optimal knife classification outcomes.
% \amin{ I am not sure if you need to talk about the experiments above. you can just say we do this because we get more information }

% How we label the knife for certain food
% \wenzhen{First talk about the method. The ground-truth labeling can be leave at the end of this subsection.}
To get the ground truth label of knife selection, we have established the following rules: 1. Use a fruit knife by default. 2. Use a serrated knife for extremely hard foods (e.g., sweet potato), foods with a hard skin (e.g., lemon), or tough and deformable foods (e.g., bread). Based on these rules, we conducted experimental cutting using a fixed policy, with details provided in Section~\ref{subsubsec: knife selection module ground truth labelling}.

% \amin{so I don't understand the last part about labeling. what are you labeling for? isn;t the rules established and selection is made?}
% \amin{ I guess you can improve the last three paragraphs. make it more consistent and  remove the repetitive information and }
% Our trial cut is designed with a fixed motion that applies a -10 N downward force and a sawing speed of 0.001 using a fruit knife. We collect data on position, velocity, and force in the z direction during this process, sampling 150 points evenly, and input them into a three-layer multi-layer perceptron (MLP). The MLP's output is a binary classification that determines whether to use a serrated or fruit knife for cutting the food.
% Once the knife is selected, our robot can switch knives easily using our specially designed two-knife mount in Fig \ref{fig: setup} (c).

% \amin{is the RL model solely for the adaptive cutting? if so, maybe you can change the title to something that mentions it}
% \amin{ I guess this section is unnecessarily long. you are overexplaining sometimes and having some unnecessary sentences like: "To effectively train RL algorithms on actual robots, it is crucial to design
% the RL settings to ensure rapid convergence"}

% \amin{or something like " at the end of each episode, the history observations, actions, and rewards are stored in the memory buffer and used for policy update" just say we use this information to update the policy or we update the policy after each cycle}
\subsection{Reinforcement learning adaptive cutting}
\label{subsec: Reinforcement learning formulation}
% \wenzhen{In general this subsection is lengthy and disorganized. Please improve the structure}

\begin{figure}[htbp]
\begin{center}
\includegraphics[scale=0.4]{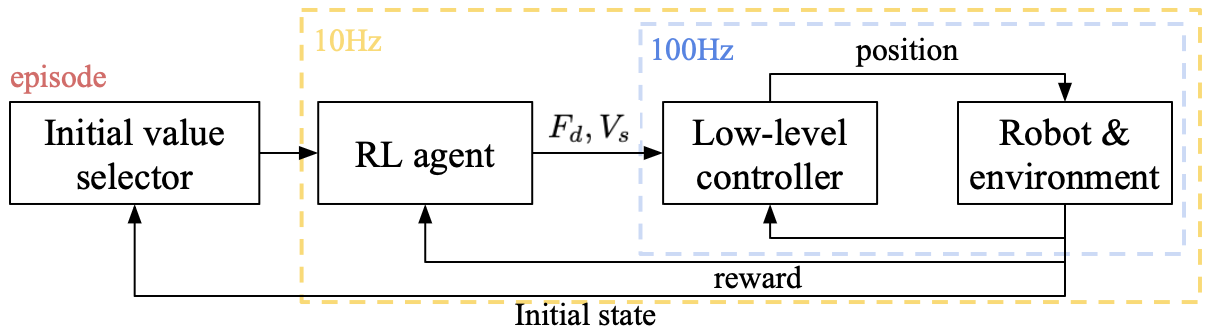}
\end{center}
\caption{\small The training pipeline of the RL agent. The low-level controller operates at $100$ Hz. The RL agent, given the reward and state, outputs the desired downward force ($F_d$) and sawing speed ($V_s$). The initial value selector runs at the beginning of each episode.}
\label{fig: rl_pipeline}
\end{figure}

\subsubsection{Reward function}
Our objective in optimizing cutting techniques is to achieve a relatively high cutting speed while minimizing the effort required to cut the food. Intuitively, using a higher desired vertical force $\widetilde{F_z}$ can accelerate the cutting process but also result in greater resistance. An appropriate sawing speed $\widetilde{V_s}$ facilitates smoother cutting but increases resistance in the sawing direction, thus consuming more energy. To find the optimal $\widetilde{F_z}$ and $\widetilde{V_s}$ that balance cutting speed and energy efficiency, we devise the following reward function:
\begin{equation}
r_t = - K_e \left( |E^x_{(t-\Delta t):t}| + |E^z_{(t-\Delta t):t}| \right) 
+ K_z \Delta z_{(t-\Delta t):t},
\end{equation}
where \( r_t \) represents the reward at time \( t \). The first term penalizes the cutting effort during the time interval \( \Delta t \), while the second term rewards downward progress of the knife. In our implementation, the energy terms are computed from the measured force and displacement along the sawing and downward directions:
\begin{align}
E^x_{(t-\Delta t):t} &= \sum_{\tau=t-\Delta t}^{t} F_x(\tau)\Delta x(\tau),\\
E^z_{(t-\Delta t):t} &= \sum_{\tau=t-\Delta t}^{t} F_z(\tau)\Delta z(\tau).
\end{align}
We use the absolute values of \(E^x_{(t-\Delta t):t}\) and \(E^z_{(t-\Delta t):t}\) because resistance in either direction contributes to the total effort required for cutting. The term \( \Delta z_{(t-\Delta t):t} \) measures the downward displacement of the knife over the same interval, incentivizing the agent to increase the cutting speed. The weights \(K_e\) and \(K_z\) were selected to normalize the energy penalty and downward-progress reward to comparable magnitudes during preliminary robot trials, so that the agent is encouraged to make progress without ignoring the energy cost.

Given this reward function, we use an on-policy RL algorithm, Proximal Policy Optimization (PPO)~\cite{ppo}, for its stable convergence, utilizing Generalized Advantage Estimation (GAE) and clipping in its actor-critic framework.

% RL intro&overview
% \wenzhen{Lots of the content here is not necessary}
% Reinforcement Learning (RL) involves an agent learning to make decisions through trial-and-error interactions with a dynamic environment. It is based on the framework of a Markov Decision Process (MDP), which consists of an observation space, action space, and a reward function. The observation space defines what the agent sees in the environment, the action space outlines possible actions the agent can take, and the reward function provides feedback based on the agent's actions, guiding it toward effective strategies to maximize cumulative rewards over time.

% Fast convergence settings
\subsubsection{Observation space and action space}
% Although most RL settings require thousands of training episodes in simulations to achieve convergence, simulating the properties of food presents significant challenges due to their unknown properties. 
To effectively train RL algorithms on actual robots, it is crucial to design the RL settings to ensure rapid convergence. Our setup features a low-dimensional observation space and a discrete action space, complemented by an initial value selector that outputs the optimal starting value to expedite agent convergence. 
% Additionally, since our RL agent is specifically designed to adjust the desired downward force $F_d$ and sawing speed $V_s$ within a limited range, rather than directly controlling the robot's terminal trajectory or joint angles, this ensures system stability.

% Observation space
The observation space is an 6-dimensional vector defined as: $O_t = [v_x, v_z, f_x, f_z, \widetilde{F_z}, \widetilde{V_s}]$.
% where \(x_r\) and \(z_r\) represent the relative displacements in the \(x\) and \(z\) axes, respectively, from the initial point of contact between the knife and the food. 
Where \(v_x\) and \(v_z\) denote the velocities of the knife in the \(x\) and \(z\) axes. \(f_x\) and \(f_z\) measure the forces exerted in the respective axes. \(\widetilde{F_z}\) and \(\widetilde{V_s}\) correspond to the desired force in the z-axis and the desired sawing speed at time \(t\), respectively.

% Action space
The action space is defined as a 4-dimensional discrete set, where each dimension corresponds to an increase and decrease of one unit in the desired force \( \widetilde{F_z} \) and the sawing speed \( \widetilde{V_s} \). Due to safety concerns and the stability of the knife mount, the target vertical force is bounded by \(-30~\mathrm{N}\leq\widetilde{F_z}\leq0~\mathrm{N}\), and the target sawing speed is constrained to \(0~\mathrm{cm/s}\leq\widetilde{V_s}\leq2~\mathrm{cm/s}\). These bounds also reduce unsafe motions and excessive deformation of soft foods during real-world RL training.

% initial value selector
\subsubsection{Initial value selector}
The initial value selector is designed to choose starting values for $\widetilde{F_z}$ and $\widetilde{V_s}$. During the initial exploration phase of training, it randomly outputs values to ensure a broad sampling of data. After collecting data for some episodes, the selector employs the critic network to choose initial values that are expected to yield the highest rewards.

% episode definition
\subsubsection{Learning pipeline}
An episode is typically defined by a single back-and-forth motion of the knife. If the duration exceeds $10$ seconds, a time-out penalty is applied, terminating the episode. Additionally, the episode also concludes if the knife contacts the cutting board. Fig \ref{fig: rl_pipeline} illustrates the pipeline of the learning process. The lower-level cutting controller, described in \ref{subsec: Low-level cutting controller}, operates at a frequency of $100$ Hz. The actor-network takes the observation as input and outputs an action at a frequency of $10$ Hz, providing \(\widetilde{F_z}\) and \(\widetilde{V_s}\) values to the lower-level controller. Thus, the commanded force and sawing speed can vary throughout a slice, although each command is tracked by the low-level controller between two consecutive policy updates. The initial value selector operates at the episode level. At the end of each episode, the history observations, actions, and rewards are stored in the memory buffer and used for policy updates.

\section{Experiment Settings}
% \wenzhen{The dataset of food should also be introduced in this section}
\label{sec: experiments}
\subsection{Experiment setup}
% \amin{ the space is messing the lines right now. double check before submission}
\begin{figure}[htbp]
\begin{center}
\includegraphics[scale=0.51]{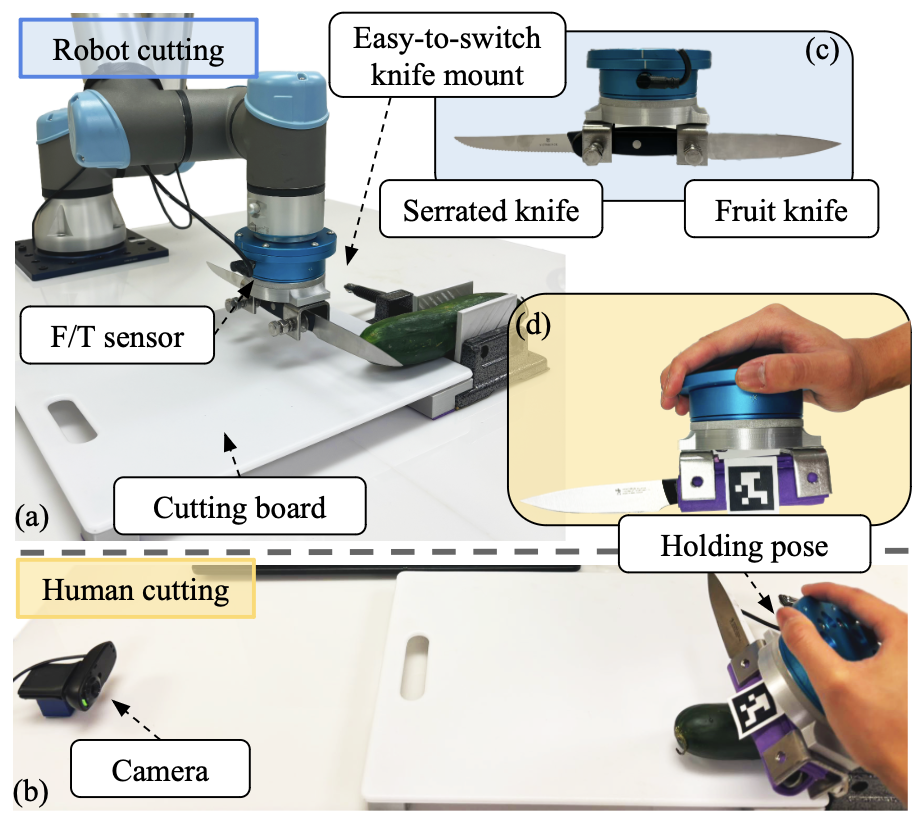}
\end{center}
\caption{\small (a) Experimental setup of the perception-manipulation robotics slicing system. (b) Experimental setup of the human evaluation system (c) Close-Up View: The easy-switch knife mount. (d) Close-Up View: Human holding the knife, ensuring accurate force/torque data collection.}
\label{fig: setup}
\end{figure}

Fig \ref{fig: setup} shows our experimental setup. Our system uses a 6-DoF UR5e Robot arm by Universal Robotics. This arm is equipped with a wrist-mounted F/T sensor from Nordbo Robotics. Fig \ref{fig: setup}(a) shows the robot setup. The easy-to-switch knife mount shown in Fig \ref{fig: setup}(c) can hold the serrated knife and fruit knife simultaneously and switch the knife by rotating the last joint of the robot by $180$ degrees without manually changing the knife. The cutting board equipped with a slide vise can hold the food tightly. 

\subsection{Human evaluation setup}
\label{sec: Human evaluation setup}

The human evaluation setup is designed to monitor force and motion during human cutting. This setup features a camera, a user interface, and a knife outfitted with a F/T sensor. The camera tracks an AprilTag \cite{apriltag} affixed to the knife, ensuring precise motion capture. When cutting, users should adopt the holding pose as depicted in Fig \ref{fig: setup}(d) to ensure the F/T sensor can accurately record the force and torque data. The human trials were used as a reference for comparison rather than as a statistically exhaustive human-subject study. Human participants were allowed to use their natural cutting styles and were not constrained by the robot's force and velocity limits.

% how does it work?
% calibration
The coefficient for converting pixel measurements to real-world distances needs calibration to ensure accurate real-world measurements. We attached an AprilTag to the knife and directed the robot to hold it while moving a known distance along the z and x axes. By measuring the corresponding pixel displacement, we determined the pixel-to-real-world conversion coefficient.

\subsection{Dataset for classification and cutting experiments}

\begin{figure}[htbp]
\begin{center}
\includegraphics[scale=0.37]{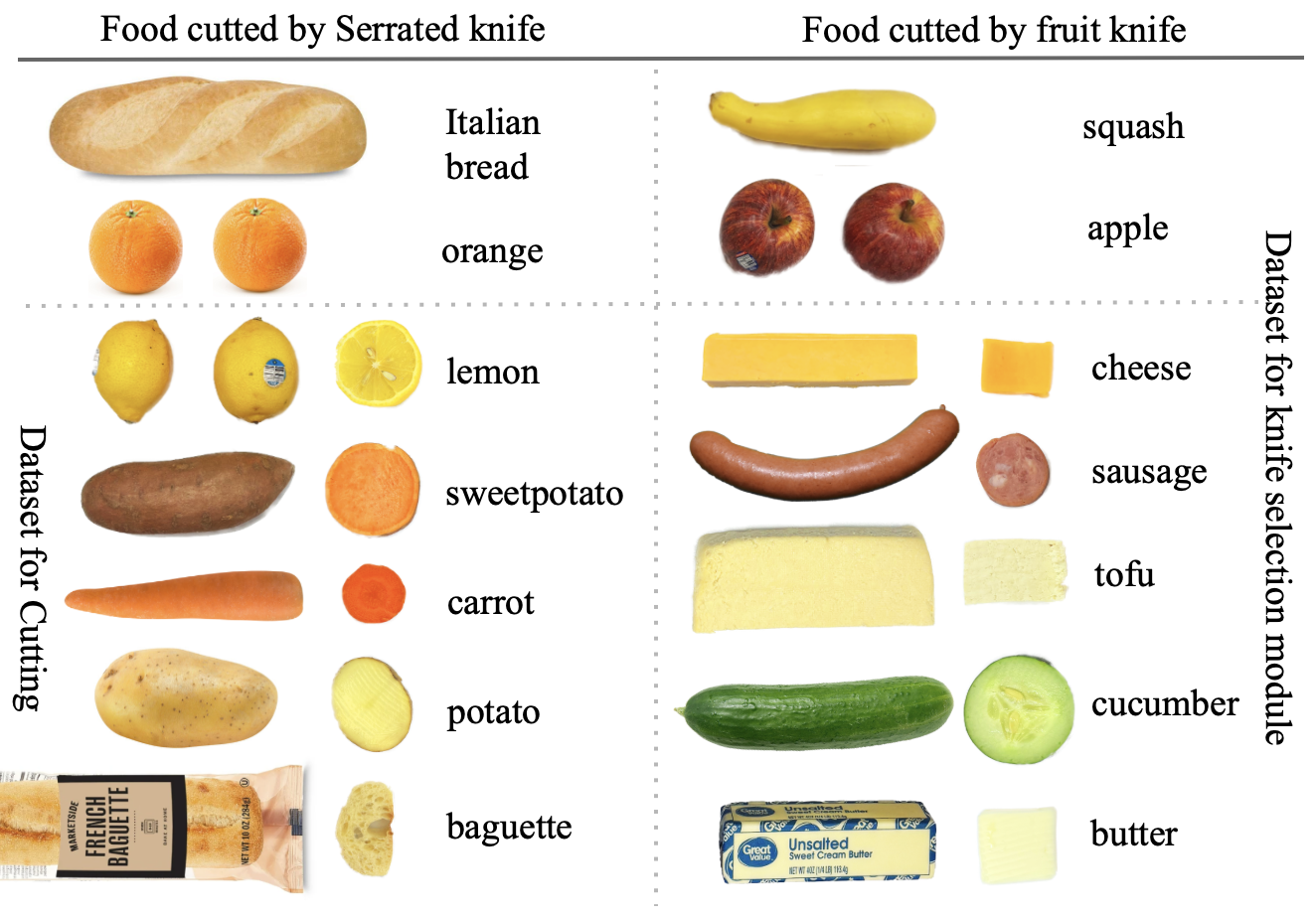}
\end{center}
\caption{\small Dataset Overview for Experiments: The left column lists foods cut with a serrated knife, while the right column lists those cut with a fruit knife. All foods serve as the dataset for the knife selection module's classification task. Additionally, ten types of food, each with one slice shown on the right, are used in cutting experiments.}
\label{fig: dataset}
\end{figure}

Fig \ref{fig: dataset} displays the variety of food used in our experiments, ranging from the softest (tofu) to the hardest (sweet potato). The left column is labeled as cutting with a serrated knife, while the right column is labeled as cutting with a fruit knife. All these foods were utilized in the knife selection module experiments. The first $2$ rows of foods served as unseen items for validation, with the remaining $10$ types used for training. Additionally, $10$ types of food, each with one slice shown on the right, are used in cutting experiments.

\subsection{Neural network and RL agent implementation}
% \wenzhen{structure relevant things should be in the method section}
The Actor network of the RL agent consists of $4$ fully connected layers, where the first $3$ layers have $128$ units each and use ReLU activation. The final layer outputs logits for action selection. The Critic network shares the structure of the Actor network but ended with a single output unit that estimates the state value. The network is updated after each episode. It is trained over $10$ epochs each episode using the Adam optimizer at a learning rate of $10^{-3}$, which decays by a factor of $0.3$ every $50$ epochs. The batch size is set at $64$. The clipping parameter is fixed at $0.2$, while the $\gamma$ and $\lambda$ parameters for the GAE are set at $0.9$ and $0.95$.

\section{Experiment results}

\label{sec: experiment result}

In this section, we first evaluate the knife selection module. Next, we train our RL adaptive controller from scratch using eight different types of food to analyze the converged policies.
% The convergence results indicate that the optimal cutting policy does not vary significantly across different foods; in fact, some fixed policies can yield quite satisfactory outcomes. 
Based on these, we propose two cutting frameworks: one with a fixed policy and the other using a pre-trained RL adaptive controller. We then compare the performance of both frameworks against human behavior.

\subsection{Knife selection module}
\label{subsec: knife selection module}
\subsubsection{Ground truth labelling}
\label{subsubsec: knife selection module ground truth labelling}
To label which knife to use with each food, we conduct the robot to perform fixed policy cutting with desired downward force as -15 N and sawing speed as 1 cm/s, which takes the middle value of our parameter range. If the fruit knife is stuck with this fixed policy, it is very likely that the fruit knife is not proper for cutting this kind of food and a serrated knife is needed, also, the learning process is likely to stuck. The labeling results are shown in Fig \ref{fig: dataset}, where the foods in the left column require a serrated knife, while those in the right column can be cut with a fruit knife.

\subsubsection{Trial cut parameter exploration}
By performing a trial cut on different foods, the knife selection module can select the proper knife for this certain food. A trial cut is conducted using a fixed target vertical force $\widetilde{F_z}$ and a sawing speed $\widetilde{V_s}$. During this process, force data are collected. The trial cut is halted once $200$ data points have been recorded. We performed multiple trial cuts with different $\widetilde{F_z}$ and $\widetilde{V_s}$ to identify the optimal policy, then combined it with a second policy to improve performance.

% We selected the $F_x$ and $F_z$ series from the dataset to characterize the food properties, as $F_z$ and $F_x$ indicate the resistance in the downward and sawing directions, respectively. We then cut the first and last $50$ data points and calculated the average $F_x$ and $F_z$ values from the middle $100$ data points as features representing the trial cut.

To optimize the parameters for the trial cut, we selected target vertical forces of $-10$ N and $-5$ N to ensure the force was not excessive for cutting through the food. For the sawing speed $\widetilde{V_s}$, we chose $0.0$, $0.5$ cm/s, and $1.0$ cm/s. We conducted $5$ trial cuts for each parameter set across $14$ different types of food, totaling $420$ trial cuts. Then we split the data into a training set and a validation set in a $3:7$ ratio and used an SVM to perform binary classification. $F_x$ and $F_z$ series are selected from the dataset to characterize the food properties, as they indicate the resistance in the vertical and sawing directions, respectively. We then trim the first and last $50$ data points and calculated the average $F_x$ and $F_z$ values from the middle $100$ data points as features representing the trial cut. The classification accuracy for each parameter set are displayed in Table \ref{tab: parameter selection}.

\begin{table}[ht]
\centering
\caption{\small Classification accuracy results of exploring different trial cut parameters.}
\label{tab: parameter selection}
\begin{tabular}{|c|c|c|}
\hline
\diagbox{$\widetilde{V_s}$}{$\widetilde{F_z}$} & $-5$ N & $-10$ N \\ \hline
$0.0$ cm/s & 0.76 & 0.71 \\ \hline
$0.5$ cm/s & 0.81 & \textbf{0.95} \\ \hline
$1.0$ cm/s & 0.90 & 0.81 \\ \hline
\end{tabular}
\end{table}

The results indicate that a target vertical force $\widetilde{F_z}$ of $-10$ N and a sawing speed $\widetilde{V_s}$ of $0.5$ cm/s most accurately reflect the food properties. To improve performance, we tried to combine it with a second set of trial cut parameters and found that of $-5$ N and $0.0$ cm/s, achieving a $100\%$ accuracy. Additionally, we tested this trial cut combination by training the SVM with $10$ types of food and testing on the remaining $4$ as an unseen validation set, achieving a success rate of $100\%$ in correctly selecting the appropriate knife.
% \amin{acually, I didn't figure out how you labeled the ground truth, the methodology part for this experiment should be improved. I have put comments there as well}
% In conclusion, we selected a combination of $2$ trial cuts with fixed parameters of $\widetilde{F_z} = -5$ N, $\widetilde{V_s}=0$ and $\widetilde{F_z} = -10$ N, $\widetilde{V_s} = 0.5$ cm/s.

\subsection{Learn the optimal cutting parameters}
\label{subsec: RL adaptive controller}

\begin{figure}[htbp]
\begin{center}
\includegraphics[scale=0.36]{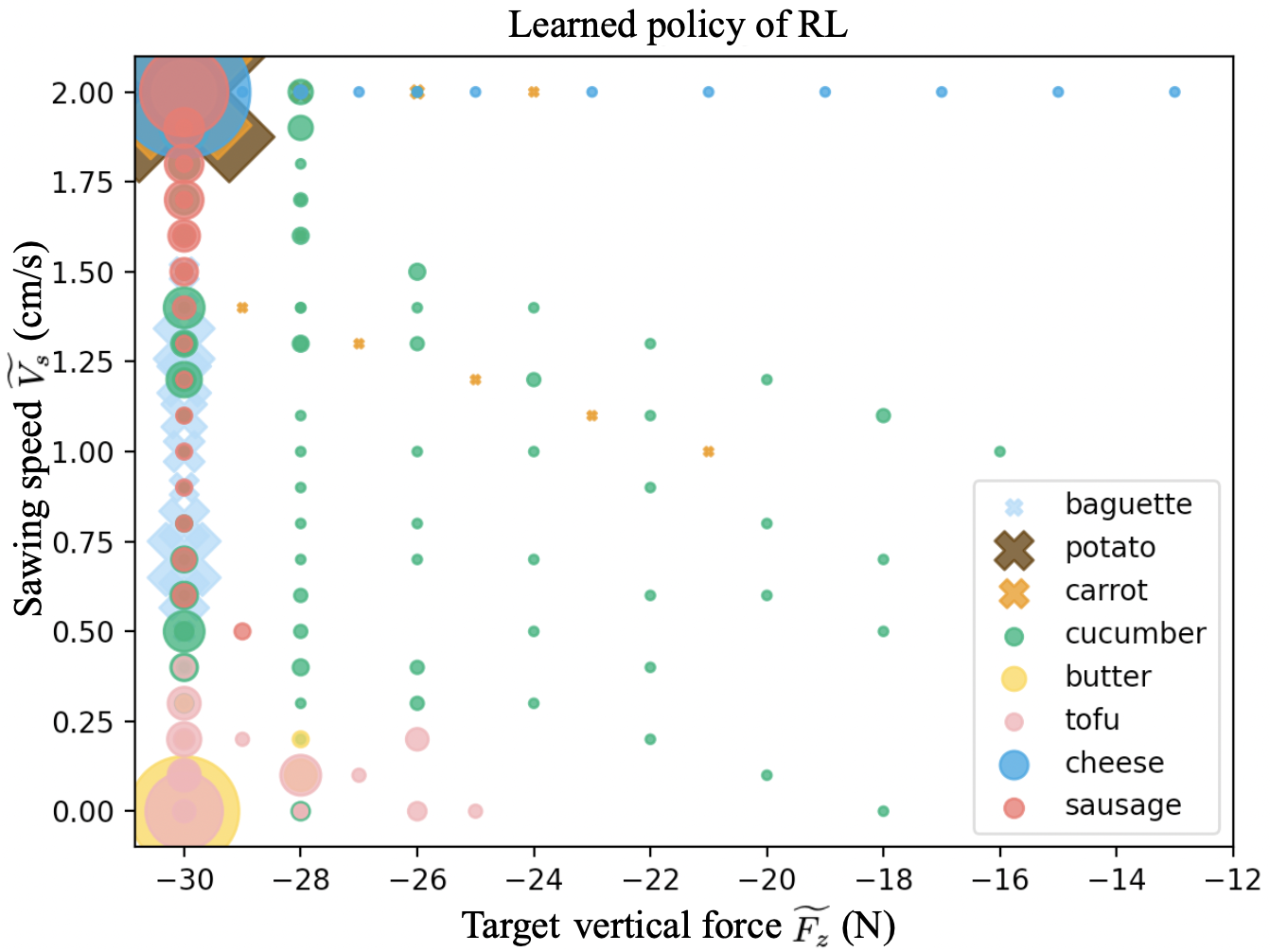}
\end{center}
\caption{\small Learned policy across different foods: the x-axis represents the desired vertical force ($\widetilde{F_z}$), and the y-axis indicates the sawing speed ($\widetilde{V_s}$) output by the actor network. Foods cut with a fruit knife are marked by circles, while those cut with a serrated knife are shown as crosses. The size of each symbol reflects the frequency with which the same policy was applied during the cutting process.}
\label{fig: learned policy}
\vspace{-3mm}
\end{figure}

% \begin{table*}[ht]  
% \centering
% \caption{Experiment result of cutting with fruit knife}
% \label{tab: fruit knife}
% \begin{tabular}{|*{16}{c|}}
% \hline
%  & \multicolumn{3}{c|}{cucumber} & \multicolumn{3}{c|}{tofu} & \multicolumn{3}{c|}{butter} & \multicolumn{3}{c|}{cheese} & \multicolumn{3}{c|}{sausage} \\ \hline 
% Eva & R & CE & CR & R & CE & CR & R & CE & CR & R & CE & CR & R & CE & CR \\ \hline
% RL & $2.27$ & 3.10 & 1.60 & 2.16 & 2.68 & 1.40 & 2.14 & 3.19 & 1.40 & 2.24 & 3.37 & 1.30 & 2.25 & 3.91 & 1.30 \\ \hline
% BS & 1.77 & 1.76 & 1.50 & 1.16 & 2.68 & 0.80 & 1.21 & 2.67 & 0.77 & 1.08 & 3.01 & 0.76 & 1.16 & 3.05 & 0.7 \\ \hline
% BC & 1.82 & 2.32 & 1.50 & 1.46 & 3.22 & 1.0 & 1.67 & 3.80 & 1.1 & 1.42 & 3.35 & 0.86 & 1.38 & 3.51 & 0.86 \\ \hline
% H & 2 & 3 & 4 & 5 & 6 & 7 & 8 & 9 & 10 & 11 & 12 & 13 & 14 & 15 & 16 \\ \hline
% \end{tabular}
% \end{table*}

\begin{table*}[ht]  
\centering
\vspace{3mm}
\caption{\small Experiment Results of Cutting. The abbreviations RL and FP represent cutting performed with the RL adaptive controller and a fixed policy, respectively; R, CE, and CR denote the evaluation metrics: Reward, Cutting Efficiency, and Cutting Rate. The first row of foods, indicated as FK at the top left, is cut using a fruit knife, while the second row, indicated as SK, is cut with a serrated knife.
Best results are highlighted in bold. For all $3$ metrics, higher values are preferable. Standard deviations are provided for results with variance from RL and human results, while they are omitted for the fixed policy results due to their relative stability.}
% \wenzhen{Labels are not clear. Please find ways to make them clear. Don't expect people to read the text carefully before looking at the table}
\label{tab: fruit knife}

\setlength{\tabcolsep}{3pt}
\begin{tabular}{|*{16}{c|}}
\hline
 FK & \multicolumn{3}{c|}{cucumber} & \multicolumn{3}{c|}{tofu} & \multicolumn{3}{c|}{butter} & \multicolumn{3}{c|}{cheese} & \multicolumn{3}{c|}{sausage} \\ \hline 
 & RL & FP & Human & RL & FP & Human & RL & FP & Human & RL & FP & Human & RL & FP & Human \\ \hline
R & $\textbf{2.4}\pm0.0$ & $1.8$ & $1.0\pm0.1$ & $\textbf{2.3}\pm0.1$ & $1.5$ & $0.9\pm0.2$ & $\textbf{2.3}\pm0.1$ & $1.7$ & $0.7\pm0.2$ & $\textbf{2.2}\pm0.1$ & $1.4$ & $0.3\pm0.2$ & $\textbf{2.3}\pm0.2$ & $1.4$ & $0.9\pm0.3$ \\ \hline
CE & $3.3\pm0.1$ & $2.3$ & $\textbf{3.5}\pm0.3$ & $2.9\pm0.1$ & $3.2$ & $\textbf{4.5}\pm0.6$ & $3.3\pm0.1$ & $\textbf{3.8}$ & $3.4\pm0.5$ & $3.3\pm0.1$ & $\textbf{3.6}$ & $2.1\pm0.5$ & $\textbf{3.8}\pm0.3$ & $3.5$ & $3.7\pm0.6$ \\ \hline
CR & $1.5\pm0.0$ & $1.5$ & $\textbf{2.6}\pm0.3$ & $1.4\pm0.2$ & $1.0$ & $\textbf{2.0}\pm0.7$ & $1.4\pm0.1$ & $1.1$ & $\textbf{1.7}\pm0.6$ & $1.0\pm0.4$ & $0.9$ & $\textbf{1.2}\pm0.5$ & $1.4\pm0.0$ & $0.9$ & $\textbf{1.7}\pm0.2$ \\ \hline \hline

SK & \multicolumn{3}{c|}{baguette} & \multicolumn{3}{c|}{potato} & \multicolumn{3}{c|}{lemon} & \multicolumn{3}{c|}{carrot} & \multicolumn{3}{c|}{sweetpotato} \\ \hline 
 & RL & FP & Human & RL & FP & Human & RL & FP & Human & RL & FP & Human & RL & FP & Human \\ \hline
R & $\textbf{2.0}\pm0.1$ & $1.2$ & $0.5\pm0.1$ & $\textbf{2.0}\pm0.0$ & $0.9$ & $1.0\pm0.1$ & $\textbf{2.0}\pm0.1$ & $1.4$ & $1.3\pm0.3$ & $\textbf{1.8}\pm0.2$ & $\textbf{1.8}$ & $0.8\pm0.1$ & $0.3\pm0.1$ & $0.2$ & $\textbf{1.0}\pm0.2$ \\ \hline
CE & $1.9\pm0.1$ & $\textbf{3.6}$ & $2.0\pm0.4$ & $\textbf{4.1}\pm0.3$ & $2.5$ & $3.5\pm0.3$ & $\textbf{3.6}\pm0.0$ & $\textbf{3.6}$ & $3.0\pm0.2$ & $2.4\pm0.1$ & $2.2$ & $\textbf{3.1}\pm0.3$ & $1.3\pm0.1$ & $1.4$ & $\textbf{2.3}\pm0.1$ \\ \hline
CR & $1.3\pm0.2$ & $0.7$& $\textbf{1.7}\pm0.0$ & $0.8\pm0.0$ & $0.7$ & $\textbf{2.2}\pm0.2$ & $1.0\pm0.0$ & $0.9 $& $\textbf{3.3}\pm1.0$ & $1.1\pm0.2$ & $1.1$ & $\textbf{1.5}\pm0.3$ & $0.7\pm0.3$ & $0.4$ & $\textbf{1.8}\pm0.4$ \\ \hline

\end{tabular}
\end{table*}
In this section, we train our RL adaptive controller from scratch using $8$ different types of food, with $5$ (butter, cheese, tofu, sausage, cucumber) cut using a fruit knife and $3$ (baguette, carrot, potato) using a serrated knife. This training was conducted to evaluate the convergence capabilities of our RL controller and to determine the final policy it adopts.

% Details of the RL settings
% To achieve a fast convergence, we simplify the observation space, assuming that the food has a uniform texture. Therefore our observation space is set as $O_t = [v_x, v_z, f_x, f_z]$, where taking the speed and force in x and z directions into consideration. And our action space is set as increase and decrease one unit in desired force $F_d$
% and desired sawing speed $V_s$. Where the limitations are set as $0N<F_d<=30N$, $0cm/s<=V_s<=20cm/s$. Therefore, the action space of the RL agent is a discrete 4-dimensional vector.

% Training process
\subsubsection{Training process}
We train the RL agent on the real robot $5$ times from scratch for each type of food. To make the RL agent broadly explore the space, we set the initial value of $\widetilde{F_z}$ and $\widetilde{V_s}$ randomly at the beginning of the first $5$ episodes. To ensure policy convergence, we monitor the loss curves of the actor and critic networks, the reward curve, and the entropy of the policy. Convergence is indicated when the loss curves stabilize, and the entropy reaches $0$. After convergence, we let the robot cut $3$ more slices with the learned policy. Typically, it takes about $20$ episodes, involving the cutting of $10$ slices, for the RL agent to converge. However, this can vary depending on the type of food and the initial conditions.

% policy comments
\subsubsection{Learned policy discussion}
Figure \ref{fig: learned policy} shows the output policy for the last $3$ slices of each food type after policy convergence. Each point on the graph represents an output policy, plotted with $\widetilde{F_z}$ on the x-axis and $\widetilde{V_s}$ on the y-axis. Circles indicate foods cut with a fruit knife, while crosses represent those cut with a serrated knife. The symbol size reflects the frequency of corresponding $\widetilde{F_z}$ and $\widetilde{V_s}$ values. Most points cluster around policies of $\widetilde{V_s}=2, \widetilde{F_z}=-30$ and $\widetilde{V_s}=0.0, \widetilde{F_z}=-30$, indicating that, for example, the controller tends to chop tofu and butter with maximum force. In contrast, it opts for a maximum-speed sawing motion for cheese. For cucumber, however, the exact sawing speed is less crucial, as convergence occurs at a moderate velocity. This behavior suggests that, under the current reward and hardware constraints, the learned policy often discovers a small number of dominant cutting modes rather than a widely varying continuous policy over the full parameter space. We therefore interpret the RL controller as an adaptive parameter-selection mechanism that can automatically choose and refine chopping/sawing behaviors from force feedback, instead of claiming that it learns a fully general continuous cutting controller.

% analysis of not achieving desired result
Why are they not converging to a value in the middle stably? Our reward function is designed to encourage finding a balance between cutting speed and energy efficiency. 
% We initially assumed that sawing reduces the force required to cut downward and this effectiveness varies with the texture of the food. Consequently, we expected to observe variations in sawing speed and the desired downward force across different foods to optimize energy efficiency. 
However, this pattern is not as distinct as expected, largely due to significant noise in the data. Another potential issue is the limitation on the desired force. While -30 N is typically sufficient, the agent consistently converges to the maximum force, suggesting it is inadequate. In contrast, the sawing speed does not always reach its maximum. However, due to safety concerns and the limitations of the robot's structure, we cannot increase the force range further. The boundary-seeking behavior also highlights a limitation of the current reward design: when faster downward progress dominates the marginal energy penalty, the optimal behavior can become biased toward the maximum allowed force.

Based on these findings, it appears that the food-cutting task can be effectively addressed using just a few fixed policies. Therefore, in the following section, we will evaluate both the RL controller and the fixed policy controller, comparing their performance to human behavior.

\subsection{Cutting result evaluation}

We evaluated the cutting result on pre-trained RL adaptive controller, fixed policy, which are baseline chopping and baseline sawing, as well as human cutting. 

\subsubsection{Rl adaptive controller (\textbf{RL})}
% setting and training of the RL controller
When deploying the RL adaptive controller for actual cutting tasks, training a policy from scratch could lead to food waste. To address this, we initially trained separate policies for fruit knives and serrated knives. The fruit knife model was trained using cucumbers, while the serrated knife model was trained on baguettes. Utilizing these pre-trained models, the RL agent can effectively cut each slice and typically converges within $1$ to $2$ slices.

\subsubsection{Fixed policy (\textbf{FP})}
We use fixed policies as our baselines, with parameters derived from the results in Section \ref{subsec: RL adaptive controller}. We choose the chopping policy $\widetilde{F_z}=-30$ N, $\widetilde{V_s}=0.0$ cm/s for the fruit knife, and sawing policy $\widetilde{F_z}=-30$ N, $\widetilde{V_s}=2$ cm/s for the serrated knife. 

\subsubsection{Human cutting results (\textbf{Human})}
Leveraging the human evaluation setup described in Section \ref{sec: Human evaluation setup}, we asked $3$ humans to cut the food using their cutting styles. Because the number of participants is small and the human trials were unconstrained, these results are intended to provide a practical reference rather than a statistically comprehensive comparison with human skill.

\subsubsection{Evaluation metrics}
To evaluate the cutting performance of the policies, we employ three metrics:
\begin{itemize}
    \item Reward (\textbf{R}): the average reward value from the last $3$ slices, as detailed in Section \ref{subsec: Reinforcement learning formulation}.
    \item Cutting Efficiency (\textbf{CE}): calculated as the average vertical cutting speed divided by the average energy consumed.
    \item Cutting Rate (\textbf{CR}): determined by dividing the food's diameter by the time taken to cut one slice.
\end{itemize}
Higher values of the three metrics indicate better performance for all metrics. Tables \ref{tab: fruit knife} shows the result. The highest values in each evaluation category are highlighted in bold. Due to the variance in RL and human results, we present both the average and the standard deviation of the last $3$ slices. For fixed policies, which exhibit less variance, only the average values are reported. The analysis shows that the RL method consistently achieves the highest reward, which is expected since it optimizes for maximum reward. RL also demonstrates comparable cutting efficiency to fixed policies and human performance, outperforming them in four instances. 

\subsubsection{Results discussion}
Fixed policies generally perform worse than RL across the three matrices, likely because RL effectively combines chopping and sawing techniques; for example, a small sawing motion can enhance chopping efficiency. However, all fixed policy cuttings were successful and produced good slices, suggesting that a correct choice of chopping or sawing policy is sufficient for basic cutting tasks, given that other experiments using a serrated knife with a chopping policy resulted in failures when cutting sweet potatoes, lemons, and carrots. This observation also explains why fixed policies are competitive in several cases: many foods can be cut by one of a few dominant modes once the knife is selected correctly. The benefit of the RL controller is that it can select and refine these modes automatically from interaction feedback, rather than relying entirely on manually chosen parameters. In contrast, humans displayed comparable cutting efficiency but did not excel in maximizing rewards. Humans consistently achieved the highest cutting rates, likely because they could use greater force and speed and were not constrained by the robot's force and velocity bounds.
% Moreover, for the most challenging food: sweet potatoes, human performance remained superior across all three metrics.

\section{Conclusion and Future Work}

% conclusion
This paper presents a perception-manipulation robotics system for slicing food. Initially, we conduct a trial cut, analyzing displacement, velocity, and force in the downward direction to determine whether to use a serrated or fruit knife. Subsequently, an RL-based adaptive controller resolves the cutting task. This controller quickly converges to an optimal policy using a pre-trained model. During training, we observed that the policy typically converges to a cutting policy utilizing either maximum downward force and zero velocity (chopping) or maximum velocity (sawing), suggesting that a fixed policy could be sufficient for the task. We assessed both the fixed policy controller (baseline) and the adaptive RL controller, comparing their performance to human cutting. The results suggest that RL is most useful in this setting as an automatic parameter-selection and refinement mechanism, while the final behaviors often correspond to interpretable chopping or sawing modes.

% future work
% imitation learning
In the future, with our developed system for evaluating human cutting, we can easily collect data and apply imitation learning to enhance the cutting task. Another important direction is to improve the reward function. The current reward mainly combines estimated mechanical work and downward progress, and therefore does not explicitly model cut quality, peak force, food deformation, viscous friction, or non-Newtonian effects that may appear in wet or soft foods such as tofu and butter. Future reward designs could include slice quality, deformation, peak-force constraints, and friction-aware terms to better capture these aspects.
% dual-arm
Moreover, we observed that the position of the holding hand is crucial for slicing, especially for deformable foods. Positioning the holding hand closer to the cutting site tends to improve cut quality. Further research is needed to develop an effective dual-arm cutting system.

\addtolength{\textheight}{0cm}   

% \section*{ACKNOWLEDGMENT}

% This work was supported by ?. The authors would like to thank the entire RoboTouch Lab for their help with this paper.

\bibliographystyle{ieeetr}
\bibliography{references.bib}

\end{document}